\pdfoutput=1
\documentclass[10pt]{article}

\usepackage[a4paper,textwidth=18cm,textheight=24cm,top=2.85cm, bottom=2.85cm, left=1.5cm, right=1.5cm]{geometry}

\usepackage{icdp2009}

\makeatletter
\long\def\@makecaption#1#2{%
\vskip\abovecaptionskip
\sbox\@tempboxa{#1. #2}%
\ifdim \wd\@tempboxa >\hsize
#1. #2\par
\else
\global \@minipagefalse
\hb@xt@\hsize{\box\@tempboxa\hfil}%
\fi
\vskip\belowcaptionskip}
\makeatother

\usepackage{lmodern}

\usepackage{graphicx}
\usepackage{times}
\usepackage{lipsum}
\usepackage{ulem}

\usepackage{svg}
\usepackage{amsmath}
\usepackage{pdfpages}
\usepackage{multirow}
\usepackage{amsmath}
\usepackage{amsfonts}
\usepackage{amssymb}
\newcommand*{\affaddr}[1]{#1} 
\newcommand*{\affmark}[1][*]{\textsuperscript{#1}}

\usepackage{hyperref}
\hypersetup{colorlinks,allcolors=black}

\usepackage{blindtext,graphicx}
\usepackage[absolute]{textpos}

\begin{document}

\noindent

\bibliographystyle{ieeetr}


\begin{textblock}{14}(1,1)
\noindent This paper is a preprint of a paper accepted by the 11th Int. Conf. on Pattern Recognition Systems (ICPRS-21)
and is subject to Institution of Engineering and Technology Copyright. When the final version is published, the copy of record will be available at the IET Digital Library.
\end{textblock}

\title{Sparse LiDAR and Stereo Fusion (SLS-Fusion) for Depth Estimation and 3D Object Detection}

\authorname{Nguyen Anh Minh Mai\affmark[1,]\affmark[2,]\affmark[$*$], Pierre Duthon\affmark[3], Louahdi Khoudour\affmark[1], Alain Crouzil\affmark[2], Sergio A. Velastin\affmark[4,]\affmark[5]}
\authoraddr{
\affaddr{\affmark[1]Cerema, Equipe-projet STI, 1 Avenue du Colonel Roche, 31400, Toulouse, France}\\
\affaddr{\affmark[2]Institut de Recherche en Informatique de Toulouse, Universit\'e de Toulouse, UPS, 31062 Toulouse, France}\\
\affaddr{\affmark[3]Cerema, Equipe-projet STI, 8-10, rue Bernard Palissy - 63017 Clermont-Ferrand Cedex 2, France}\\
\affaddr{\affmark[4]Department of Computer Science and Engineering, Universidad Carlos III de Madrid, \\28911 Leganés, Madrid, Spain}\\
\affaddr{\affmark[5]School of Electronic Engineering and Computer Science, Queen Mary University of London, UK}\\
\affaddr{\affmark[]$*$ nguyen-anh-minh.mai@cerema.fr}}



\maketitle


\abstract

The ability to accurately detect and localize objects is recognized as being the most important for the perception of self-driving cars. From 2D to 3D object detection, the most difficult is to determine the distance from the ego-vehicle to objects. Expensive technology like LiDAR can provide a precise and accurate depth information, so most studies have tended to focus on this sensor showing a performance gap between LiDAR-based methods and camera-based methods.
Although many authors have investigated how to fuse LiDAR with RGB cameras, as far as we know there are no studies to fuse LiDAR and stereo in a deep neural network for the 3D object detection task. This paper presents SLS-Fusion, a new approach to fuse data from 4-beam LiDAR and a stereo camera via a neural network for depth estimation to achieve better dense depth maps and thereby improves 3D object detection performance. 
Since 4-beam LiDAR is cheaper than the well-known 64-beam LiDAR, this approach is also classified as a low-cost sensors-based method. 
Through evaluation on the KITTI benchmark, it is shown that the proposed method significantly improves depth estimation performance compared to a baseline method. Also, when applying it to 3D object detection, a new state of the art on low-cost sensor based method is achieved.

\section{Introduction} 






Accurately detecting obstacles in their surroundings is key for safe driving of an autonomous vehicle. The current top-performance methods in 3D object detection \cite{shi2020pv, he2020structure, shi2019pointrcnn} are mainly based on LiDAR technology. Alternatively, systems based only on camera sensors also receive much attention because of their low cost and wide range of use \cite{chabot2017deep, you2019pseudo, wang2019pseudo, li2019stereo, qin2019triangulation, chen2020dsgn, li2020confidence}
. Even though much research has focused on camera-based methods, the gap in performance between monocular or stereo-based methods and LiDAR-based methods is still significant, meriting more research.


The work reported here aims at designing a 3D object detection architecture from low-cost sensors because we are interested in the practical applications of using them in self-driving vehicles like an autonomous bus. 
There are many 3D object detection architectures that will be presented in the next section.
In recent years, a new interesting and promising branch of research proposed by You \textit{et al.} \cite{you2019pseudo}, namely Pseudo-LiDAR, generates pseudo point clouds based on the estimated depth map from images using the pinhole camera model. Then, they can be treated as a LiDAR signal input for any LiDAR-based 3D object detector. This method has shown a significant improvement in performance compared to previous image-based object detection methods with this simple conversion.

Real LiDAR sensors can provide depth information with high accuracy in the form of 3D points. However, these are very sparse point clouds depending on the properties of the objects in the scene. Hence, point density is normally uneven, and it causes difficulties for recognizing obstacles. Furthermore, in denser point clouds obtained by the Pseudo-LiDAR pipeline \cite{wang2019pseudo}, distant or small objects such as pedestrians and cyclists, for which LiDARs usually generate fewer points, can get more points and so be more easily detected and localized if the predicted depth map from monocular or stereo-based method is good enough. Finally, another advantage of Pseudo-LiDAR is that it can combine the state of the art of depth estimation and of 3D object detection. Therefore, the performance for 3D object detection of this kind of methods based on this idea strongly depends on the estimated depth map.
Instead of using expensive 64 laser beams which costs about \$75,000, You \textit{et al.} \cite{you2019pseudo} simulate the 125 times cheaper 4-beam LiDAR Scala by sparsifying the original 64-beam signals and use these points to correct their estimated depth map (by Stereo Depth Net (SDN)) by applying a graph-based depth correction (GDC).

\begin{figure}[t!]
    \centering
    \includegraphics[width=0.45\textwidth]{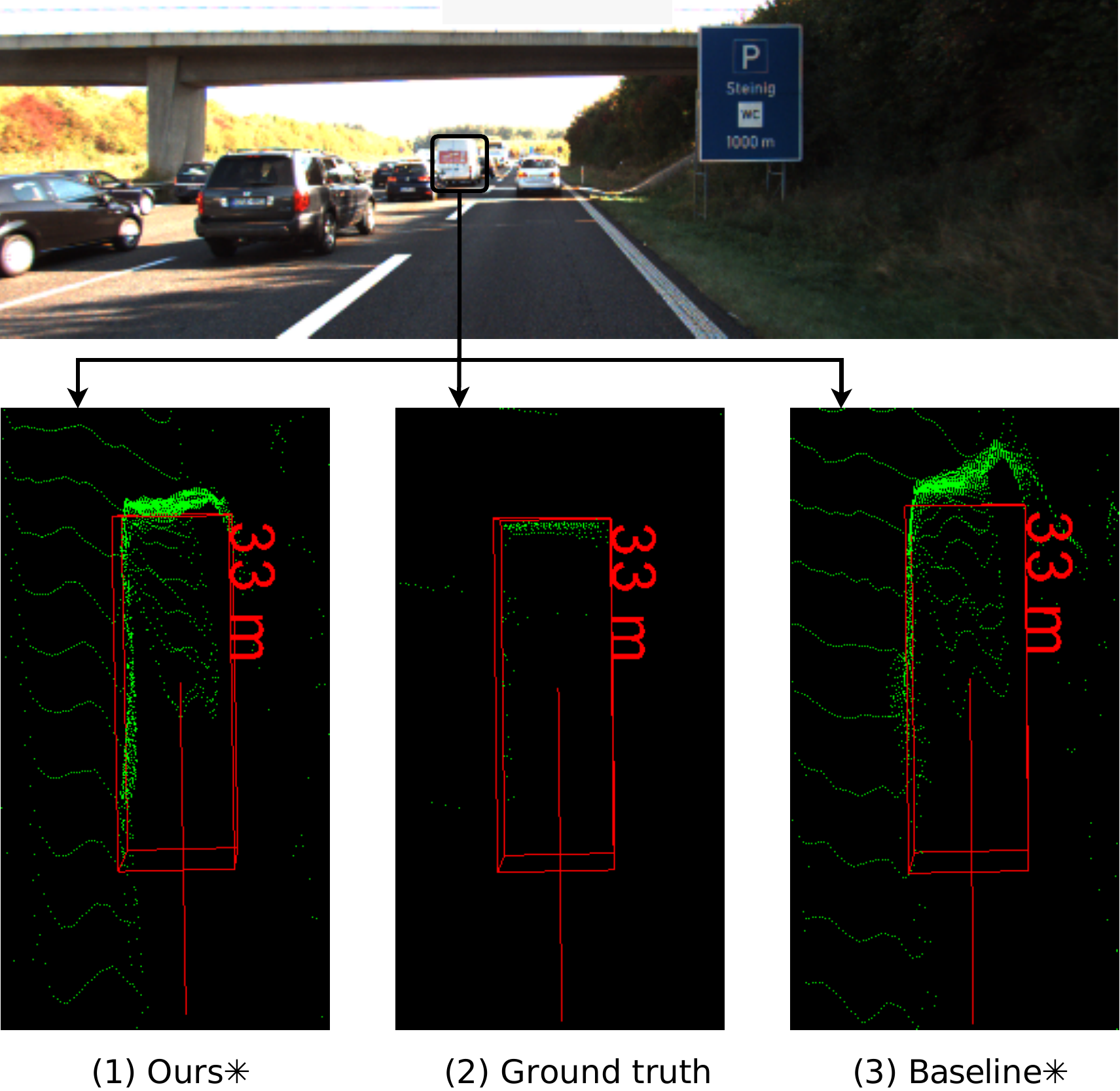}
    \caption{An example of pseudo point cloud generated by the proposed depth network and the baseline Pseudo-LiDAR++ (SDN) \cite{you2019pseudo}. The figure shows the 3D point cloud (\textcolor{green}{green}) from bird’s-eye view (BEV) representation corresponding to a car at 33 m in the top picture in the KITTI dataset. They are respectively point cloud from the proposed method (denoted by Ours$*$) (1), ground truth (2) and the Pseudo-LiDAR++ (SDN) (denoted by Baseline$*$) (3). The 3D bounding box (\textcolor{red}{red}) denotes the ground truth location of the car.} 
    \label{fig:1}
\end{figure}

Starting from this work, a neural network is proposed here that takes 4-beam LiDAR and stereo images as input. The new method estimates a dense depth map and then uses it for the 3D object detection task.
Unlike \cite{you2019pseudo}, taking stereo images to estimate the depth map and then integrating the 4-beam LiDAR as a post-processing step to correct the estimated depth map, this approach fuses the 4-beam LiDAR with stereo images in deep neural network to obtain a more accurate depth map.
Fig. \ref{fig:1} illustrates that when using this method, the resulting pseudo point cloud is closer to the ground truth than the baseline Pseudo-LiDAR++ (SDN) method \cite{you2019pseudo}.

To summarize, the main contributions in this paper are as follows:
\begin{itemize}  

    \item A Sparse LiDAR and Stereo Fusion network (SLS-Fusion) is proposed, which is the first method fusing LiDAR and stereo together in a deep neural network to focus on 3D object detection. 
    \item By integrating the GDC proposed in \cite{you2019pseudo} with the proposed SLS-Fusion, experimental results on the validation KITTI object detection dataset demonstrate that the proposed approach can produce very accurate depth maps thanks to LiDAR-Stereo fusion. This outperforms all previous low-cost based methods for 3D object detection task.
\end{itemize}

\section{Related work}

\begin{figure*}[!ht]
    \centering

    \includegraphics[width=0.99\textwidth]{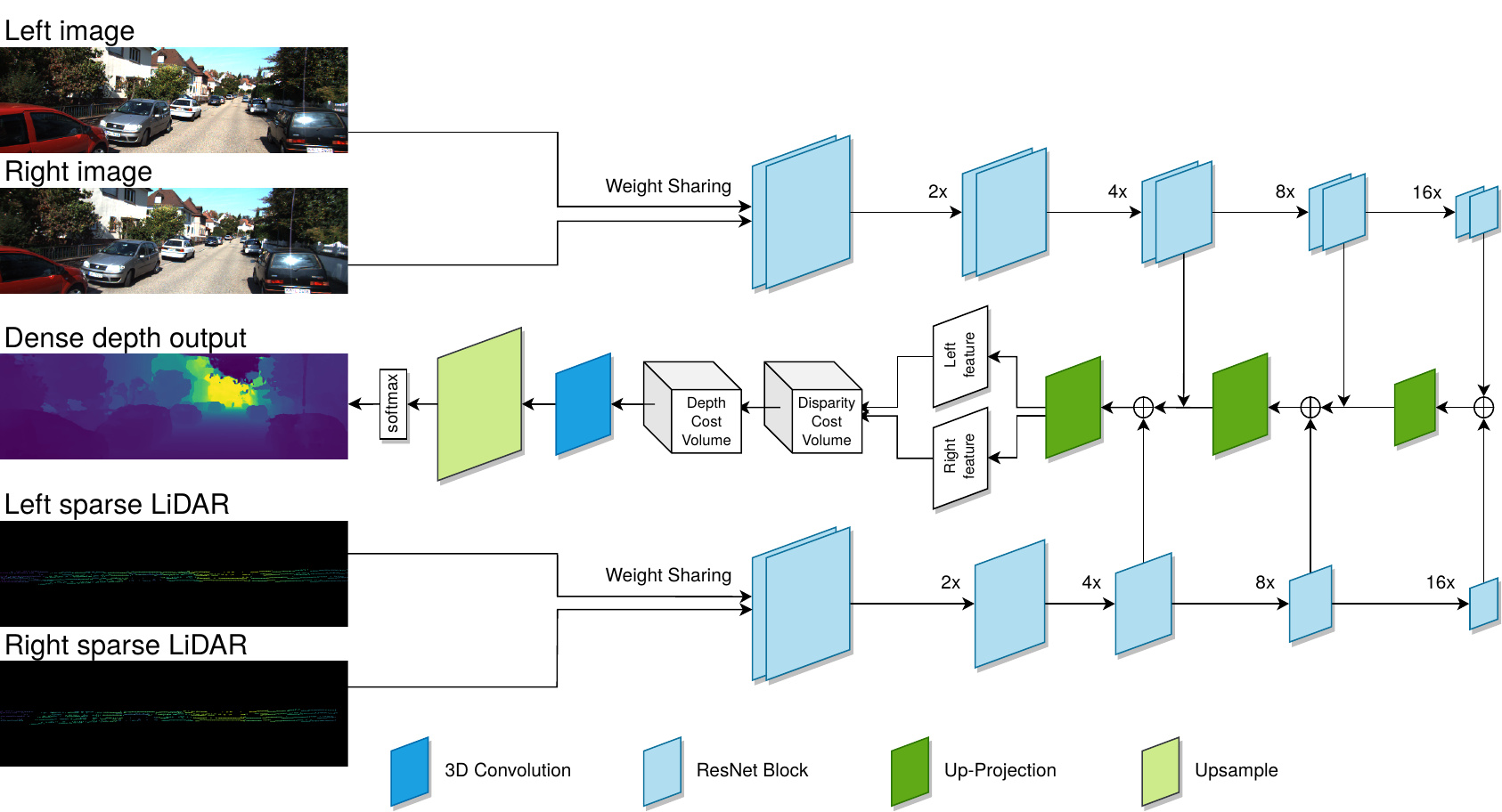}
    
    \caption{\textbf{The pipeline of SLS-Fusion}. The left and right input stereo images, $I_{l}$ and $I_{r}$, are fed to two weight-sharing pipelines consisting of ResNet blocks to encode features shown on the top flow (the same way for the left and right projected 4-beam LiDAR images, $S_{l}$ and $S_{r}$, on the bottom flow). In the decoding phase, each LiDAR and image pair are fused, ($S_{l}$, $I_{l}$) and ($S_{r}$, $I_{r}$) and corresponding feature tensors are obtained, left feature and right feature, for each LiDAR-image input pair. Then left and right feature tensors go through  Depth Cost Volume (DeCV) proposed in \cite{you2019pseudo} to directly learn the depth.}
    \label{fig: depthnet}
\end{figure*}


\noindent \textbf{Depth Completion.} 
Depth completion aims to predict a denser depth map from image and sparse depth map.
Qiu \textit{et al.} \cite{qiu2019deeplidar} proposed the first study extending the surface normal as guidance to estimate depth map for outdoor environments. They combined a single RGB image, LiDAR point cloud and the surface normal to produce a dense depth map, but their network is very large (around 144 million parameters), and therefore time-consuming needing a powerful architecture for training.


\noindent \textbf{3D Object Detection.} Most 3D object detection methods  \cite{shi2020pv, he2020structure, shi2019pointrcnn} achieving good performance are strongly based on LiDAR exploiting its precision. 
Meanwhile, methods based only on camera sensors \cite{chabot2017deep, you2019pseudo, wang2019pseudo, li2019stereo, qin2019triangulation, li2020confidence, pon2019object} face difficulties to get precise depth. Chatbot \textit{et al.} \cite{chabot2017deep} presented one of the first studies in monocular-based 3D object detection. It is mainly based on detecting 2D keypoints and finding the best position through 2D and 3D matching. However, it is time-consuming and requires a 3D model dataset including several types of vehicles. A recent study by Wang \textit{et al.} \cite{wang2019pseudo} showed surprising results by detecting objects on pseudo point clouds. Some other methods \cite{qi2018frustum, wang2019frustum} try to fuse LiDAR point cloud and image. Qi \textit{et al.} \cite{qi2018frustum} proposed to detect 2D bounding boxes from a 2D image and further extract corresponding 3D bounding frustums. Finally, they predict 3D bounding boxes on interior points. However, its performance heavily relies on the 2D object detector which may fail due to some drawbacks of the camera such as illumination or ill-posed problem. 
Sensor fusion is a hot topic but how to effectively fuse these sensors is difficult and still needs more research.



\noindent \textbf{From 2D to 3D object detection.} Detecting and localizing objects in 3D from 2D is challenging. Existing approaches usually generate several proposal boxes from an image or a bird's eye view projection and then predict the 3D bounding box. Recently, a new research direction has been proposed in \cite{wang2019pseudo} which detects objects on pseudo LiDAR generated from the estimated depth map by using a simple conversion. It achieved good results and inspired many other studies \cite{you2019pseudo, li2020confidence, pon2019object, wang2020task}. You \textit{et
al.} \cite{you2019pseudo} proposed SDN network with Depth Cost Volume to improve depth map accuracy and integrated a depth correction step to regularize the pseudo point cloud. Wang \textit{et al.} \cite{wang2020task} studied the effect of foreground and background pixels, and they proposed a foreground-background separation for depth estimation. 
\cite{li2020confidence, pon2019object} improved the accuracy of the predicted depth map by integrating semantics from RGB images.



\noindent \textbf{LiDAR-Stereo Fusion.} There have been very few attempts to fuse LiDAR and stereo in autonomous driving. Previous methods \cite{cheng2019noise, park2018high, wang20193d} were designed for the depth completion and, as far as we know, there is no fusion involved for 3D object detection benchmark dataset. 
Cheng \textit{et al.} \cite{cheng2019noise} reported existing noise in LiDAR signals such as misalignment between LiDAR and stereo cameras. They then proposed an unsupervised learning method to predict the dense depth map by fusing LiDAR and stereo which guides noise filtering in LiDAR point cloud. 
Wang \textit{et al.} \cite{wang20193d} extracted features from LiDAR and stereo images and proposed the Conditional Cost Volume Normalization. The main aim is to adaptively regularize cost volume optimization based on LiDAR signals.

The approach proposed here is specifically intended for the depth completion task and tested on a 3D object detection task.
Both Pseudo-LiDAR++ (SDN) \cite{you2019pseudo} and the proposed SLS-Fusion method use Depth Cost Volume (DeCV) \cite{you2019pseudo} to directly minimize the depth error instead of the disparity error. Thereby, we range differences can be avoided between depth $D$ and disparity $d$ value for each pixel $(u, v)$ expressed through $D(u, v) = \frac{f_{U} \times b}{d(u, v)}$, where $f_{U}$ and b are respectively the horizontal focal length and the baseline in the stereo system. Unlike \cite{you2019pseudo}, a new backbone is designed that enables to take left, right RGB images and the image projection of LiDAR. This paper presents a new approach which fuses stereo images and 4-beam LiDAR and predict a dense depth map.

\section{Method}
The goal is to detect and localize 3D bounding box of objects by using stereo RGB images and 4-beam LiDAR. 
Given left and right sparse depth maps $S_{l}$, $S_{r}$ generated by projecting the sparse LiDAR point cloud on the left and right image planes using the calibration parameters and the left and right RGB image $I_{l}$, $I_{r}$, the Sparse Lidar and Stereo Fusion network estimates the dense depth map $D$ of the whole image. The overall pipeline of the depth network is shown in Fig. \ref{fig: depthnet}.
\subsection{Motivation}
Wang \textit{et al.} \cite{wang2019pseudo} presented Pseudo LiDAR which improves the performance of the previous 3D object detection methods based on RGB/D image by changing the nature of the input. You \textit{et al.}  \cite{you2019pseudo} created their own depth estimation network to improve the accuracy of the depth map and thereby to increase the 3D object detection performance. We think that adding LiDAR, which allows to have strong and accurate depth information into the network to have more features, can improve the performance of the model. The inexpensive 4-beam LiDAR system, is integrated instead of using the well-known but expensive 64-beam LiDAR. This hypothesis has led to the design of the SLS-Fusion network.
\subsection{Input}
To enrich the representation for a normal stereo (RGBs) matching network, a decision has been taken to join the geometry  information from the LiDAR point cloud.
However, instead of directly using a 3D point cloud from LiDAR, like in \cite{wang20193d},  the 4-beam LiDAR point cloud is reprojected to both left and right image coordinates using the calibration parameters to obtain the two sparse 4-beam LiDAR depth maps corresponding to stereo images. Unlike \cite{wang20193d} that used a simple early fusion paradigm by concating stereo images with their corresponding sparse LiDAR depth maps, the proposed method uses a late fusion approach, presented in the following section.
\subsection{SLS-Fusion Network}
The model can be divided into two parts: feature extraction and Depth Cost Volume \cite{you2019pseudo}. 



\noindent \textbf{Feature extraction.} Qiu \textit{et al.} \cite{qiu2019deeplidar} proposed a Deep Completion Unit which is an encoder-decoder network taking a late fusion strategy, where extracted features from RGB images and LiDAR point cloud are combined in the decoding phase. Inspired by this work, a network architecture is proposed here that combines LiDAR and image in a late fusion strategy. To further leverage the stereo input, as in \cite{you2019pseudo, chang2018pyramid}, a weight-sharing pipeline is used for both LiDAR and image, ($I_{l}$, $S_{l}$) and ($I_{r}$ , $S_{r}$), instead of left and right images. The encoder consists of a series of ResNet blocks followed by a convolution layer with a stride of 2 to downsize the feature maps by a ratio of 1/16 of input to obtain more detailed features for small objects in the image. To upsize the feature maps and to integrate features from both encoders for LiDAR and image, an up-projection layer is used \cite{laina2016deeper} for the decoder. However, unlike \cite{qiu2019deeplidar}, only 3 up-projection layers are used to obtain a tensor for a balance between feature resolution (1/4 of input) and the number of feature channels to later put into the Depth Cost Volume, which is computationally expensive.

\noindent \textbf{Depth Cost Volume.} The left and right features obtained from the decoding stage are passed to the Depth Cost Volume (DeCV) proposed by You \textit{et al.} \cite{you2019pseudo} to learn the depth. In fact, the two features are fed to Disparity Cost Volume (DiCV) \cite{chang2018pyramid} to form a 4D cost tensor, and then it is converted to DeCV to directly optimize the distance loss instead of the disparity loss. The smooth L1 loss function from \cite{you2019pseudo} is used.
\section{Experiments}
\subsection{Experimental Setup}
\noindent \textbf{Datasets.} Datasets are divided into 2 parts: A) for depth estimation and B) for 3D object detection.

A) For depth estimation, we take advantage of the diversity of a large scale dataset, Scene Flow dataset \cite{mayer2016large}, which contains 35,454 training and 4,370 testing images of size $960\times 540$ from an array of synthetic scenes. This is a large scale synthetic dataset which provides dense and elaborate disparity maps as ground truth to train the proposed network. 

\cite{wang2019pseudo, weng2019monocular} use the training data from KITTI depth completion dataset which overlaps with KITTI object detection validation part and induces overfitting. 
Meanwhile, \cite{you2019pseudo, wang2019pseudo} use the LiDAR 64 beams as depth map ground truth for training their depth network. Instead, we use the KOD dataset provided by Wang \textit{et al.} \cite{wang2020task}, which provides the relation between the KITTI object detection dataset and the KITTI depth completion dataset. So, for the ground truth, instead of using LiDAR 64 beams depth maps, we have denser depth maps from KITTI depth completion dataset which combines 11 frames of LiDAR.

B) For 3D object detection, the proposed approach is evaluated on the KITTI 3D object detection benchmark \cite{geiger2013vision}, which is the most well known dataset for self-driving cars. It contains 7,481 training samples and 7,518 testing samples. LiDAR point clouds, RGB images and also camera calibration matrices are provided as input. Following Chen \textit{et al.} \cite{chen2017multi}, the training dataset is split into a training part (3,712 samples) and a validation part (3,769 samples). The benchmark has annotations for 3 classes, which are cars, pedestrians and cyclists. 



\noindent \textbf{Evaluation metrics.} For the evaluation of object detection, we report average precision (AP) with Intersection over Union (IoU) thresholds at 0.5 and 0.7, as per normal practice.  We denote AP for the 3D and bird's eye view (BEV) tasks as $AP_{3D}$ and $AP_{BEV}$, respectively. Object detection is divided into three levels of difficulty: easy, moderate and hard, depending on the 2D bounding box sizes, occlusion, and truncation extent following the KITTI definitions. Like \cite{li2020confidence}, experiments are compared to six recent stereo-based detectors: TLNet \cite{qin2019triangulation}, Stereo-RCNN \cite{li2019stereo}, 
DSGN \cite{chen2020dsgn}, CG-Stereo: PointRCNN \cite{li2020confidence}, Pseudo-LiDAR: PointRCNN \cite{wang2019pseudo}, the baseline method Pseudo-LiDAR++: PointRCNN \cite{you2019pseudo} and also the original LiDAR-based detector PointRCNN \cite{shi2019pointrcnn}. But unlike \cite{li2020confidence}, we don't compare with the result of OC-Stereo \cite{pon2019object} which used the AVOD \cite{ku2018joint} detector while the proposed method uses the PointRCNN \cite{shi2019pointrcnn} detector.



Results are also reported to evaluate the depth network. Following the KITTI depth completion benchmark, the root mean squared error [mm] (RMSE) is used as the main metrics. Root mean squared error of the inverse depth [1/km] (iRMSE), mean absolute error [mm] (MAE) and mean absolute error of the inverse depth [1/km] (iMAE) are also reported.
A comparison is made with the Stereo Depth Network (SDN), which is the depth network of Pseudo-LiDAR++ (SDN) (You \textit{et al.} \cite{you2019pseudo}), to test the advantage of LiDAR feature in the deep network.  



\begin{table*}[ht!]
\begin{center}
\begin{tabular}{|l|c|c|c|c|c|c|c|}
\hline
\multicolumn{1}{|c|}{\multirow{2}{*}{Method}} & \multirow{2}{*}{Input} & \multicolumn{3}{c|}{0.5 IoU} & \multicolumn{3}{c|}{0.7 IoU} \\\cline{3-8}
& & Easy & Moderate & Hard & Easy & \textbf{\textcolor{red}{Moderate}} & Hard \\
\hline\hline
TLNet \cite{qin2019triangulation} & S & 62.46 / 59.51 & 45.99 / 43.71 & 41.92 / 37.99 & 29.22 / 18.15 & 21.88 / 14.26 & 18.83 / 13.72 \\
Stereo-RCNN \cite{li2019stereo} & S & 87.13 / 85.84 & 74.11 / 66.28 & 58.93 / 57.24 & 68.50 / 54.11 & 48.30 / 36.69 & 41.47 / 31.07 \\
Pseudo LiDAR \cite{wang2019pseudo} & S & 88.4 / 88.0 & 76.6 / 73.7 & 69.0 / 67.8 & 73.4 / 62.3 & 56.0 / 44.9 & 52.7 / 41.6\\
DSGN \cite{chen2020dsgn} & S & - & - & - & 83.24 / 73.2 & 63.91 / 54.27 & 57.83 / 47.71\\
CG-Stereo \cite{li2020confidence} & S & \textbf{97.04} / 90.58 & 88.58/ \textbf{87.01} & 80.34/ 79.76 & 87.31 / 76.17 & 68.69 / 57.82 & 65.80 / 54.63 \\
\hline\hline
 
\textcolor{blue}{Baseline$*$ \cite{you2019pseudo}} & \textcolor{blue}{S} & \textcolor{blue}{89.8} / \textcolor{blue}{89.7} & \textcolor{blue}{83.8} / \textcolor{blue}{78.6} & \textcolor{blue}{77.5} / \textcolor{blue}{75.1} & \textcolor{blue}{82.0} / \textcolor{blue}{67.9} & \textcolor{blue}{64.0} / \textcolor{blue}{50.1} & \textcolor{blue}{57.3} / \textcolor{blue}{45.3} \\%
\textcolor{blue}{Ours$*$} & \textcolor{blue}{L4+S} & \textbf{\textcolor{blue}{90.13}} / \textbf{\textcolor{blue}{89.9}} & \textbf{\textcolor{blue}{83.89}} / \textbf{\textcolor{blue}{78.81}} & \textbf{\textcolor{blue}{78.03}} / \textbf{\textcolor{blue}{76.14}} & \textbf{\textcolor{blue}{82.38}} / \textbf{\textcolor{blue}{68.08}} & \textbf{\textcolor{blue}{65.42}} / \textbf{\textcolor{blue}{50.81}} & \textbf{\textcolor{blue}{57.81}} / \textbf{\textcolor{blue}{46.07}}\\%
\hline\hline
Baseline$**$ \cite{you2019pseudo}& L4+S & 90.3 / 90.3 & 87.7 / \textbf{86.9} & \textbf{84.6} / \textbf{84.2} & \textbf{88.2} / 75.1 & \textbf{76.9} / 63.8 & 73.4 / \textbf{57.4} \\%
Ours$**$ & L4+S & \textbf{93.16} / \textbf{93.02} & \textbf{88.81} / 86.19 & 83.35 / 84.02 & 87.51 / \textbf{76.67} & 76.88 / \textbf{63.90} & \textbf{73.55} / 56.78 \\%
\hline\hline

\textcolor{gray}{PointRCNN \cite{shi2019pointrcnn}} & \textcolor{gray}{L64} & \textcolor{gray}{97.3 / 97.3} & \textcolor{gray}{89.9 / 89.8} & \textcolor{gray}{89.4 / 89.3} & \textcolor{gray}{90.2 / 89.2} & \textcolor{gray}{87.9 / 78.9} & \textcolor{gray}{85.5 / 77.9} \\

\hline
\end{tabular}
\end{center}

\caption{$AP_{BEV}$/ $AP_{3D}$ results on KITTI validation set for "Car" category with IoU at 0.5 and 0.7 and on three levels of difficulty: Easy, Moderate and Hard. The results are evaluated using the original KITTI metric with 11 recall positions. All methods are ranked based on the moderately difficult $AP_{3D}$ results (red).  S, L4 and L64 respectively denote stereo, simulated 4-beam LiDAR and 64-beam LiDAR inputs.
Baseline$*$, Baseline$**$, Ours$*$ and Ours$**$ respectively are Pseudo-LiDAR++ (stereo), Pseudo-LiDAR++ (stereo + LiDAR 4 beam), the proposed method SLS-Fusion (stereo) and SLS-Fusion (stereo + LiDAR 4 beam).
}
\label{table:od}
\end{table*}

\noindent \textbf{Implementation Details.} The procedure used is quite similar to the common pipeline of Pseudo-LiDAR. 

For the depth estimation, the depth network is implemented in Pytorch and its backbone is inspired from that of DeepLiDAR. For faster convergence, full depth maps from Scene Flow dataset are firstly used for training. Here the LiDAR input is set as zeros because this synthetic dataset is designed exclusively for RGB stereo and no LiDAR scans are provided. Then fine-tuning is done on the 3,712 training samples of the KOD dataset for 100 epochs, with batch size 4 and the learning rate set to 0.001. The Adam optimizer was used to optimize the network on 2 NVIDIA GeForce GTX 1080 Ti GPUs with 11G memory. You \textit{et al.} \cite{you2019pseudo} is followed to generate the simulated 4-beam LiDAR (as similar as possible to 4-beam LiDAR ScaLa sensor) from 64-beam LiDAR and then project it on the left and right image planes to feed them into the depth network.


Once the predicted depth map $D$ is obtained, with $D(u, v)$ being the depth corresponding to each image pixel $(u, v)$, a pseudo point cloud can be generated by using the pinhole camera model. Given the depth estimate and camera intrinsic matrix, deriving the 3D location ($X_{c}$, $Y_{c}$, $Z_{c}$) in the camera coordinate system for each pixel $(u, v)$ is simply as:
\begin{subequations}\label{eqn:abc}
    \begin{align}
        \textrm{(depth) } Z_{c} &= D(u,v)\label{eqn:abc1}\\
        \textrm{(width) } X_{c} &= \frac{(u-c_{U})\times Z_{c}}{f_{U}}\label{eqn:abc2}\\
        \textrm{(height) } Y_{c} &= \frac{(v-c_{V})\times Z_{c}}{f_{V}}\label{eqn:abc3}
    \end{align}
\end{subequations}
where $c_{U}$ and $c_{V}$ is the pixel location corresponding to the camera center; $f_{U}$ and $f_{V}$ is the horizontal and the vertical focal length. Then this point is transformed into ($X_{l}$, $Y_{l}$, $Z_{l}$) in the LiDAR coordinate system (the real world coordinate system). Given the camera extrinsic matrix $C=\begin{bmatrix}
 R&t \\ 
 0&1 
\end{bmatrix}$, where $R$ and $t$ are respectively the rotation matrix and translation vector. The pseudo point clouds can be obtained by computing:
\begin{align}
    \begin{bmatrix}
    X_{l}\\
    Y_{l}\\ 
    Z_{l}\\
    1
    \end{bmatrix}
    = C^{-1}\begin{bmatrix}
    X_{c}\\
    Y_{c}\\ 
    Z_{c}\\
    1
    \end{bmatrix}
\end{align}
\noindent To make pseudo point cloud more similar to real LiDAR signals, following \cite{wang2019pseudo, yang2018pixor},  reflectance is set to 1 for each point and points higher than 1m are removed. Afterwards, LiDAR-based 3D object detector can be applied on these points. 

For 3D object detector, PointRCNN \cite{shi2019pointrcnn} is applied, which is a LiDAR-based method with a high performance and used by many other methods as a baseline. As it was designed to take into account the sparse point cloud, the dense point cloud is sub-sampled to 64 lines LiDAR. Then, the released implementations of PointRCNN is used directly, and their guidelines followed to train it on the training set of KITTI object detection dataset only for the ``Car'' class because car is one of the main objects and occupies the largest percentage in KITTI dataset which causes the unbalance between ``Car'' and other classes.






\begin{table}[ht!]
\begin{center}
\begin{tabular}{|l|c|c|c|c|}
\hline
\multicolumn{1}{|c|}{\multirow{2}{*}{Method}} & \textbf{\textcolor{red}{RMSE}} & MAE & iRMSE  & iMAE\\
& (mm) & (mm) & (1/km) & (1/km)\\
\hline\hline
Baseline$*$ \cite{you2019pseudo} & 1506.92 & 443.81 & 5.59 & 1.90 \\
Ours$*$ & \textbf{845.21} & \textbf{307.36} & \textbf{2.72} & \textbf{1.28} \\

\hline
\end{tabular}
\end{center}
\caption{Study on depth estimation. The results show the mean error (shown in different metrics) for a pixel from all test images of KOD dataset \cite{wang2020task}. RMSE (red) is the main metric to rank methods.}
\label{table:depth}
\end{table}

\subsection{Experimental Results}

This section firstly presents an evaluation of the depth estimation accuracy of the SLS-Fusion network. The results are shown in Tab. \ref{table:depth}. 
The results show that the proposed network outperforms the baseline Pseudo-LiDAR++ (SDN) \cite{you2019pseudo} by a large margin over all metrics evaluated on pixels with a valid depth ground truth in the range [1 m, 80 m] on the KOD dataset \cite{wang2020task}. 
They support the idea that adding LiDAR in stereo network can boost the accuracy of the predicted depth map.

Following the promising results from the predicted map, results for 3D object detection task are reported in Tab. \ref{table:od}, showing two results for the proposed approach. Ours$*$ is the base proposed model, while Ours$**$ includes the GDC proposed in \cite{you2019pseudo}. Pseudo-LiDAR++ is taken as a baseline and therefore in the same way, Baseline$*$ and Baseline$**$ denote respectively Pseudo-LiDAR++ (SDN) and Pseudo-LiDAR++ (SDN+GDC). The proposed approach is also compared with six other outstanding methods: Pseudo LiDAR \cite{wang2019pseudo}, TLNet \cite{qin2019triangulation}, Stereo-RCNN \cite{li2019stereo}, 
DSGN  \cite{chen2020dsgn}, CG-Stereo \cite{li2020confidence} which are based on low-cost sensors and the original 3D object detector PointRCNN \cite{shi2019pointrcnn} which uses 64-beam LiDAR. 
Firstly, comparing Ours$*$ with the Baseline$*$, the proposed method outperforms the baseline over all indicators, thereby showing the effect of 4-beam LiDAR in the proposed stereo matching model. 
Secondly, Ours$**$ outperforms the Baseline$**$ and other stereo-based methods on some indicators.
The various results can be explained by the cumbersome of the model or the training way.
The proposed method gets  the best result (63.90\%) on 3D AP with IoU=0.7 which is the main indicator to rank all methods on KITTI.
Lastly, compared to the original detector PointRCNN \cite{shi2019pointrcnn} which used 64-beam LiDAR as input, Ours$*$ and Ours$**$ have inferior performance, which is expected, but it helps to get closer to the method trained on real LiDAR.

The good results achieved by CG-Stereo \cite{li2020confidence} on 0.5 IoU (easy and moderate) could be explained by the method used. In fact, this latter splits the scene into foreground/ background and applies on each part independent depth estimation. This leads to easier detection of "easy" objects in the foreground. However, we notice that our method is more stable because it achieves better results on hard objects.

\section{Conclusion}
This paper has introduced SLS-Fusion network, a novel depth prediction framework taking 4-beam LiDAR and stereo images as input which are low-cost sensors in the autonomous driving domain, for 3D object detection.
To the best of our knowledge, there are no studies to fuse LiDAR and stereo together in a deep neural network which focus on the 3D object detection task in the literature.
In contrast to existing fusion-based method in other tasks, often found in indoor scenes, a network is proposed that can take advantage of features from both LiDAR and images by using an encoder-decoder network and then optimize the depth loss via a Depth Cost Volume \cite{you2019pseudo} for the road scenes.
Experimental results on the validation KITTI dataset demonstrate that this model can improve both depth estimation and 3D object detection accuracy.


\noindent \textbf{Perspectives and Future work.} 
We plan to train and submit our results on testing part (without ground truth) of KITTI object detection dataset to have a more objective assessment about this approach. We also evaluate results on other classes such as pedestrian or cyclist and to study the density of pseudo point cloud instead of sub-sampling pseudo point cloud like in PointRCNN detector \cite{shi2019pointrcnn}.

\bibliography{icdp2009}

\begin{thebibliography}{10}

\bibitem{shi2020pv}
S.~Shi, C.~Guo, L.~Jiang, Z.~Wang, J.~Shi, X.~Wang, and H.~Li, ``{PV-RCNN}:
  Point-voxel feature set abstraction for 3{D} object detection,'' in {\em
  CVPR}, pp.~10529--10538, 2020.

\bibitem{he2020structure}
C.~He, H.~Zeng, J.~Huang, X.-S. Hua, and L.~Zhang, ``Structure aware
  single-stage 3{D} object detection from point cloud,'' in {\em CVPR},
  pp.~11873--11882, 2020.

\bibitem{shi2019pointrcnn}
S.~Shi, X.~Wang, and H.~Li, ``{PointRCNN}: 3{D} object proposal generation and
  detection from point cloud,'' in {\em CVPR}, pp.~770--779, 2019.

\bibitem{chabot2017deep}
F.~Chabot, M.~Chaouch, J.~Rabarisoa, C.~Teuliere, and T.~Chateau, ``{Deep
  MANTA}: A coarse-to-fine many-task network for joint 2{D} and 3{D} vehicle
  analysis from monocular image,'' in {\em CVPR}, pp.~2040--2049, 2017.

\bibitem{you2019pseudo}
Y.~You, Y.~Wang, W.-L. Chao, D.~Garg, G.~Pleiss, B.~Hariharan, M.~Campbell, and
  K.~Q. Weinberger, ``Pseudo-{L}i{DAR}++: Accurate depth for 3{D} object
  detection in autonomous driving,'' {\em arXiv preprint arXiv:1906.06310},
  2019.

\bibitem{wang2019pseudo}
Y.~Wang, W.-L. Chao, D.~Garg, B.~Hariharan, M.~Campbell, and K.~Q. Weinberger,
  ``Pseudo-{L}i{DAR} from visual depth estimation: Bridging the gap in 3{D}
  object detection for autonomous driving,'' in {\em CVPR}, pp.~8445--8453,
  2019.

\bibitem{li2019stereo}
P.~Li, X.~Chen, and S.~Shen, ``Stereo {R-CNN} based 3{D} object detection for
  autonomous driving,'' in {\em CVPR}, pp.~7644--7652, 2019.

\bibitem{qin2019triangulation}
Z.~Qin, J.~Wang, and Y.~Lu, ``Triangulation learning network: from monocular to
  stereo 3{D} object detection,'' in {\em CVPR}, pp.~7607--7615, 2019.

\bibitem{chen2020dsgn}
Y.~Chen, S.~Liu, X.~Shen, and J.~Jia, ``{DSGN}: Deep stereo geometry network
  for 3{D} object detection,'' in {\em CVPR}, pp.~12536--12545, 2020.

\bibitem{li2020confidence}
C.~Li, J.~Ku, and S.~L. Waslander, ``Confidence guided stereo 3{D} object
  detection with split depth estimation,'' {\em arXiv preprint
  arXiv:2003.05505}, 2020.

\bibitem{qiu2019deeplidar}
J.~Qiu, Z.~Cui, Y.~Zhang, X.~Zhang, S.~Liu, B.~Zeng, and M.~Pollefeys,
  ``{DeepLiDAR}: Deep surface normal guided depth prediction for outdoor scene
  from sparse {LiDAR} data and single color image,'' in {\em CVPR},
  pp.~3313--3322, 2019.

\bibitem{pon2019object}
A.~D. Pon, J.~Ku, C.~Li, and S.~L. Waslander, ``Object-centric stereo matching
  for 3{D} object detection,'' {\em arXiv preprint arXiv:1909.07566}, 2019.

\bibitem{qi2018frustum}
C.~R. Qi, W.~Liu, C.~Wu, H.~Su, and L.~J. Guibas, ``Frustum {PointNets} for
  3{D} object detection from {RGB-D} data,'' in {\em CVPR}, pp.~918--927, 2018.

\bibitem{wang2019frustum}
Z.~Wang and K.~Jia, ``{Frustum ConvNet}: Sliding frustums to aggregate local
  point-wise features for amodal 3{D} object detection,'' {\em arXiv preprint
  arXiv:1903.01864}, 2019.

\bibitem{wang2020task}
X.~Wang, W.~Yin, T.~Kong, Y.~Jiang, L.~Li, and C.~Shen, ``Task-aware monocular
  depth estimation for 3{D} object detection.,'' in {\em AAAI},
  pp.~12257--12264, 2020.

\bibitem{cheng2019noise}
X.~Cheng, Y.~Zhong, Y.~Dai, P.~Ji, and H.~Li, ``Noise-aware unsupervised deep
  {LiDAR}-stereo fusion,'' in {\em CVPR}, pp.~6339--6348, 2019.

\bibitem{park2018high}
K.~Park, S.~Kim, and K.~Sohn, ``High-precision depth estimation with the 3{D}
  {LiDAR} and stereo fusion,'' in {\em ICRA}, pp.~2156--2163, 2018.

\bibitem{wang20193d}
T.-H. Wang, H.-N. Hu, C.~H. Lin, Y.-H. Tsai, W.-C. Chiu, and M.~Sun, ``{3D
  LiDAR} and stereo fusion using stereo matching network with conditional cost
  volume normalization,'' {\em arXiv preprint arXiv:1904.02917}, 2019.

\bibitem{chang2018pyramid}
J.-R. Chang and Y.-S. Chen, ``Pyramid stereo matching network,'' in {\em CVPR},
  pp.~5410--5418, 2018.

\bibitem{laina2016deeper}
I.~Laina, C.~Rupprecht, V.~Belagiannis, F.~Tombari, and N.~Navab, ``Deeper
  depth prediction with fully convolutional residual networks,'' in {\em 3DV},
  pp.~239--248, 2016.

\bibitem{mayer2016large}
N.~Mayer, E.~Ilg, P.~Hausser, P.~Fischer, D.~Cremers, A.~Dosovitskiy, and
  T.~Brox, ``A large dataset to train convolutional networks for disparity,
  optical flow, and scene flow estimation,'' in {\em CVPR}, pp.~4040--4048,
  2016.

\bibitem{weng2019monocular}
X.~Weng and K.~Kitani, ``Monocular 3{D} object detection with {Pseudo-LiDAR}
  point cloud,'' in {\em ICCV Workshops}, 2019.

\bibitem{geiger2013vision}
A.~Geiger, P.~Lenz, C.~Stiller, and R.~Urtasun, ``Vision meets robotics: The
  {KITTI} dataset,'' {\em The International Journal of Robotics Research},
  vol.~32, no.~11, pp.~1231--1237, 2013.

\bibitem{chen2017multi}
X.~Chen, H.~Ma, J.~Wan, B.~Li, and T.~Xia, ``Multi-view 3{D} object detection
  network for autonomous driving,'' in {\em CVPR}, pp.~1907--1915, 2017.

\bibitem{ku2018joint}
J.~Ku, M.~Mozifian, J.~Lee, A.~Harakeh, and S.~L. Waslander, ``Joint 3{D}
  proposal generation and object detection from view aggregation,'' in {\em
  IROS}, pp.~1--8, IEEE, 2018.

\bibitem{yang2018pixor}
B.~Yang, W.~Luo, and R.~Urtasun, ``Pixor: Real-time 3d object detection from
  point clouds,'' in {\em Proceedings of the IEEE conference on Computer Vision
  and Pattern Recognition}, pp.~7652--7660, 2018.

\end{thebibliography}

\end{document}